%% file: main.tex
\documentclass[journal]{IEEEtran}

\IEEEoverridecommandlockouts                              

\usepackage{graphicx}
\usepackage{epstopdf}
\usepackage{subcaption}
\usepackage{color}
\usepackage{multirow}
\usepackage{adjustbox}
\usepackage{booktabs}
\usepackage[table,xcdraw]{xcolor}
\usepackage{caption}
\usepackage{amsmath}
\usepackage{url}
\usepackage{array} 
\usepackage{graphicx}
\usepackage{amsmath}
\usepackage{amssymb}
\usepackage{booktabs}
\usepackage{cite}
\usepackage{bbding}

\definecolor{others}{rgb}{0, 0, 0}
\definecolor{barrier}{rgb}{1, 0.47058824, 0.19607843}
\definecolor{bicycle}{rgb}{1, 0.75294118, 0.79607843}
\definecolor{bus}{rgb}{1, 1, 0.0}
\definecolor{car}{rgb}{0.0, 0.58823529, 0.96078431}
\definecolor{construction}{rgb}{0, 1, 1}
\definecolor{motorcycle}{rgb}{1, 0.49803922, 0}
\definecolor{pedestrian}{rgb}{1, 0, 0}
\definecolor{cone}{rgb}{1, 0.94117647, 0.58823529}
\definecolor{trailer}{rgb}{0.52941176, 0.23529412, 0}
\definecolor{truck}{rgb}{0.62745098, 0.1254902, 0.94117647}

\definecolor{driveable}{rgb}{1, 0, 1}
\definecolor{flat}{rgb}{0.54509804,0.5372549,0.5372549}
\definecolor{sidewalk}{rgb}{0.29411765,0,0.29411765}
\definecolor{terrain}{rgb}{0.58823529,0.94117647,0.31372549}
\definecolor{manmade}{rgb}{0.90196078,0.90196078,0.98039216}
\definecolor{vegetation}{rgb}{0,0.68627451,0}
\definecolor{ego_vehicle}{rgb}{0,0,0}

\title{\LARGE \bf
Inverse++: Vision-Centric 3D Semantic Occupancy Prediction Assisted with 3D Object Detection
}

\author{Zhenxing Ming, Julie Stephany Berrio,  Mao Shan, Stewart Worrall
\thanks{The authors are with the Australian Centre for Robotics (ACFR) at the University of Sydney (NSW, Australia). E-mails: {\tt\small{\{d.ming, j.berrio, m.shan,
s.worrall}\}@acfr.usyd.edu.au}.}%
}

\begin{document}


\twocolumn[{%
 \renewcommand\twocolumn[1][]{#1}
 \maketitle
 \input{teaser_fig}
 }]

\thispagestyle{empty}
\pagestyle{empty}

\begin{abstract}
3D semantic occupancy prediction aims to forecast detailed geometric and semantic information of the surrounding environment for autonomous vehicles (AVs) using onboard surround-view cameras.  Existing methods primarily focus on intricate inner structure module designs to improve model performance, such as efficient feature sampling and aggregation processes or intermediate feature representation formats. In this paper, we explore multitask learning by introducing an additional 3D supervision signal by incorporating an additional 3D object detection auxiliary branch. This extra 3D supervision signal enhances the model's overall performance by strengthening the capability of the intermediate features to capture small dynamic objects in the scene, and these small dynamic objects often include vulnerable road users, i.e. bicycles, motorcycles, and pedestrians, whose detection is crucial for ensuring driving safety in autonomous vehicles. Extensive experiments conducted on the nuScenes datasets, including challenging rainy and nighttime scenarios, showcase that our approach attains state-of-the-art results, achieving an IoU score of 31.73\% and a mIoU score of 20.91\% and excels at detecting vulnerable road users (VRU). The code will be made available at: \url{https://github.com/DanielMing123/Inverse++}
\end{abstract}

\begin{IEEEkeywords}
autonomous vehicles, 3D semantic occupancy prediction, environment perception
\end{IEEEkeywords}

\section{INTRODUCTION}
Understanding the three-dimensional (3D) geometry of the surrounding environment is a fundamental aspect in the advancement of autonomous vehicle (AV) systems to guarantee safety. In recent years, vision-centric AV systems have gained significant attention as a promising approach due to their cost-effectiveness, stability, and versatility. This approach takes advantage of surround view images as input and has demonstrated competitive performance in various 3D perception tasks, including depth estimation \cite{surrounddepth, r3d3}, 3D object detection \cite{bevdepth, bevformer, bevformerv2}, 3D object tracking \cite{uatrack,startrack}, and online high definition (HD) map generation \cite{maptr, maptrv2, hdmapnet}. The introduction of 3D semantic occupancy prediction, involving voxelizing 3D space and assigning occupancy probabilities to each voxel, has further improved the 3D perception capabilities of AVs. We assert that 3D semantic occupancy serves as a suitable representation of the vehicle's surrounding environment. This representation inherently ensures geometric consistency and accurately describes occluded areas. Moreover, it exhibits greater robustness towards object classes that are not present in the training dataset. Researchers in the field have explored various techniques \cite{tpvformer,surroundocc,occformer} to predict the 3D semantic occupancy of a scene. However, although these methods have potential, their reliance on only a single 3D supervision signal (Fig.\ref{teaser}, single supervision approach) or a single 3D supervision signal combined with an additional 2D supervision signal (Fig.\ref{teaser}, dual supervision approach) for model training may cause the failure to capture small dynamic objects effectively due to lacking extra 3D training signal that forces the model to pay attention to those objects. In particular, such objects frequently include vulnerable road users (VRU), including bicycles, motorcycles, and pedestrians.

To address the aforementioned limitation and enhance the model's ability to capture VRU, we propose a method called Inverse++ (Fig.\ref{teaser}, our approach). In this approach, we introduce an additional 3D object detection auxiliary branch to the main branch. This auxiliary branch provides extra 3D supervision signals, which directly affect the intermediate features of the model. The purpose of these additional 3D supervision signals is to prioritize the model's attention towards small dynamic objects on the road. Through comparisons with other state-of-the-art (SOTA) algorithms on the nuScenes dataset, including challenging rainy and nighttime scenarios, we demonstrate that our method not only excels in its overall SOTA performance but also achieves the best performance in detecting VRU related classes, i.e. pedestrians, motorcycles, and bicycles, which is a critical task for autonomous driving and road safety. 

This paper represents a substantial expansion of our previous work, InverseMatrixVT3D \cite{ming2024inversematrixvt3d}, which focuses on a vision-only approach for the prediction of 3D semantic occupancy. The main contributions of this paper are outlined as follows: 
\begin{itemize}
    \item We propose Inverse++, a novel vision-centric 3D semantic occupancy prediction framework that utilizes an additional 3D object detection auxiliary branch to enhance performance and achieve superior results.
    \item We introduce a query-based 3D object detection auxiliary branch that provides an additional 3D supervision signal to effectively supervise the intermediate features in the main branch.
    \item We compare our approach with other state-of-the-art (SOTA) algorithms in the 3D semantic occupancy prediction task to prove the effectiveness of our method.
\end{itemize}

The remainder of this paper is structured as follows: Section \ref{literature} provides an overview of related research and identifies the key differences between this study and previous publications. Section \ref{model} outlines the general framework of Inverse++ and offers a detailed explanation of the implementation of each module. Section \ref{simulation} presents the results of our experiments. Finally, Section \ref{conclusion} provides the conclusion of our work.

\section{Related Work}\label{literature}

\begin{figure*}[h]
     \centering
     \begin{subfigure}[]{\textwidth}
         \centering
         \includegraphics[width=\textwidth]{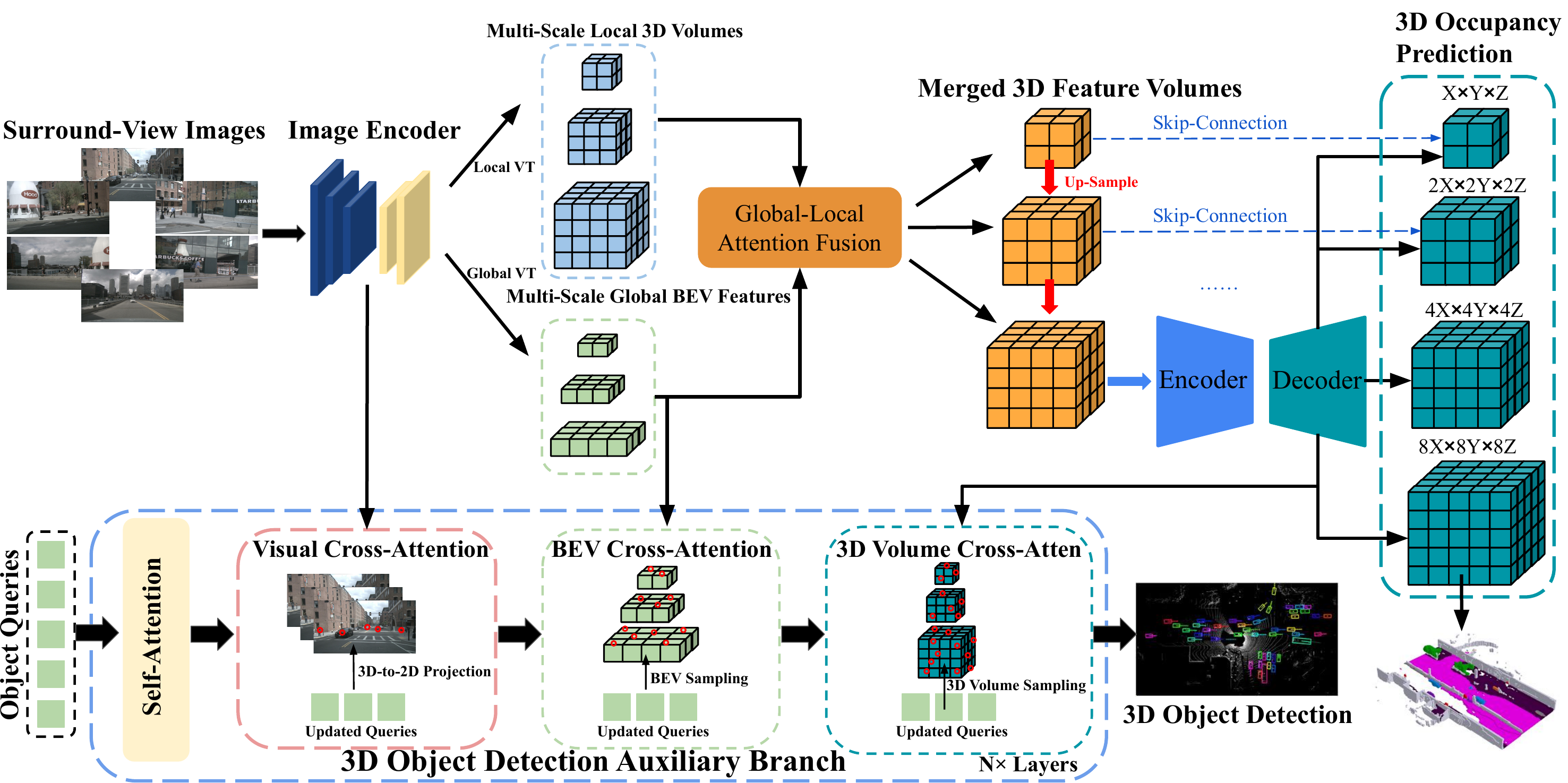}
     \end{subfigure}
        \caption{\small \textbf{Overall architecture of Inverse++.} The pipeline comprises two branches: the main branch includes an image encoder for extracting multi-scale visual features, global and local view transformations to produce intermediate multi-scale global BEV features and 3D feature volumes, global-local attention fusion to yield merged multi-scale 3D feature volumes, and a UNet-like Encoder-Decoder structure for further feature refinement, culminating in the final multi-scale 3D volume logits. The 3D object detection auxiliary branch introduces an extra 3D supervision signal that applies to visual features, multi-scale global BEV features, and multi-scale 3D volume logits. This auxiliary branch enhances the model's capability to effectively capture small dynamic objects.}
        \label{Inverse++}
\end{figure*}

\subsection{Single 3D Supervision Signal Based 3D Semantic Occupancy Prediction}
Based on the success of bird's eye view (BEV) perception algorithms \cite{LSS,bevdet,bevdepth,bevstereo,bevheight,bevdet4d,bevformer,bevformerv2,bevfusion}, several works \cite{milo, multi, tpvformer, occformer, radocc,geocc,octreeocc,surroundocc, openoccupancy,panoocc,daocc,sparseocc} have advanced the development of perception algorithms to do 3D modeling regarding the surround scenes of AVs. These methods aim to construct 3D feature volume from surround-view visual features and then feed it to a specific head to perform the 3D semantic occupancy prediction task. These approaches rely on a single 3D supervision signal and improve model performance through the implementation of enhanced view transformation techniques and carefully integrated feature refinement modules.

Despite achieving impressive performance, these methods overlook the critical aspect of integrating additional 3D supervision signals for model training. Solely depending on a single 3D supervision signal impairs the model's generalizability and may lead to training bias towards specific categories due to imbalanced instances within each class. Our study presents a novel approach that addresses the aforementioned constraints by incorporating a 3D object detection auxiliary branch. This inclusion introduces supplementary 3D supervision signals during the training phase, enhancing the model's ability to detect small dynamic objects, often essential vulnerable road users on the streets.

\subsection{3D+2D Dual Supervision Signal Based 3D Semantic Occupancy Prediction}
In the research conducted by \cite{fbocc}, the author enhances the model's generality by introducing an additional 2D semantic segmentation branch to offer extra 2D supervision signals. Additionally, in order to harness the potential of big data, the author employs the SAM algorithm \cite{sam} to generate a considerable amount of ground truth for city-driving scenarios to maximize the availability of 2D supervision signals. The 2D semantic segmentation auxiliary branch exclusively utilizes surround-view visual features as input and trains these features for 2D semantic segmentation to enhance their semantic comprehension. Simultaneously, the surround-view visual features are directed to downstream structures for 3D semantic occupancy prediction. The resulting model performance sees significant improvement attributed to the additional 2D supervision signal and the utilization of big data.

While this approach achieved remarkable performance, they overlook the limitation of extra 2D semantic segmentation signals. The additional 2D supervision signal provides significantly less information for heavily occluded objects, leading the model to prioritize foreground objects and struggle to accurately detect heavily occluded background objects. In contrast, our method utilizes supplementary 3D supervision signals from the 3D object detection task, adept at effectively handling heavily occluded objects.


\subsection{3D+2.5D Dual Supervision Signal Based 3D Semantic Occupancy Prediction}
In the study by \cite{fastocc}, the authors utilize the view transformation method introduced in \cite{LSS} in conjunction with a 2D CNN-based Encoder-Decoder structure to derive final BEV features. They then employ a similar structure proposed in \cite{flashocc} for the task of 3D semantic occupancy prediction. Concurrently, they integrate an additional 2.5D BEV segmentation auxiliary branch into the primary model branch. This auxiliary branch introduces an extra 2.5D supervision signal, applied to the final BEV features for simultaneous BEV segmentation. This enhancement elevates model performance in managing partially obscured objects, improves the detection of background elements like buildings, sidewalks, and drivable surfaces, and expands the model's perceptual scope.

The addition of a 2.5D BEV segmentation branch enhances the model's ability to handle occluded objects. However, the performance improvement is constrained as the additional 2.5D supervision signal, which lacks height information, leads to a degradation in detection accuracy. Moreover, the imbalance in instances within the 2.5D supervision signal introduces bias, directing the model's focus more towards background static objects such as buildings and drivable surfaces than foreground dynamic objects like cars, buses, and motorcycles.

\section{Inverse++}\label{model}
In this study, our main objective is to generate a dense 3D semantic occupancy grid of the surrounding scene using surround-view images ($I = \{Img^{1}, Img^{2}, \cdots, Img^{N}\}$). Additionally, we aim to improve the final 3D occupancy grid by introducing an additional 3D object detection supervision signal. Thus, the problem at hand can be described in the following manner:
\begin{equation}
    Occ_{\_}logits,BEV = NN(Img^{1},Img^{2},\dots,Img^{N})
\end{equation}
\begin{equation}
    Objects = Aux_{\_}NN(Occ_{\_}logits, BEV)
\end{equation}
\begin{equation}
    Occ = MLP(Occ_{\_}logits)
\end{equation}
where $NN$ is the neural network that utilizes view transformations to aggregate visual features and obtain the final 3D volume logits and BEV features. The $Aux_{\_}NN$, on the other hand, refers to the auxiliary branch for 3D object detection, which takes the 3D volume logits and BEV features as input. By optimizing the 3D object detection task through the training process, the capability of the 3D volume logits and BEV features in capturing small dynamic objects is enhanced. The final results of the prediction of 3D semantic occupancy can be obtained by inputting the logits of the 3D volume into a multilayer perceptron (MLP). It is denoted as $Occ \in \mathbb{R}^{X \times Y \times Z}$ and represents the semantic property of the grids, with values ranging from 0 to 16. In our case, a class value of 0 indicates that the grid is empty.

\subsection{Overview}
Fig. \ref{Inverse++} shows the overall architecture of our method. Given a set of surround view images, we use an image encoder consisting of a 2D backbone and neck to extract $N$ cameras and $L$ levels of multiscale visual features $V=\left \{ \left \{ V_{n}^{l} \right \}_{n=1}^{N}\in \mathbb{R}^{C_{l} \times H_{l} \times W_{l}} \right \} _{l=1}^{L}$. Then, both global and local view transformations proposed in \cite{ming2024inversematrixvt3d} are applied to multiscale visual features $V$ to obtain multiscale local 3D volumes $Occ_{xyz}^{l}\in \mathbb{R}^{C_{l}\times X_{l}\times Y_{l} \times Z_{l}}$ and BEV features $BEV_{xy}^{l}\in \mathbb{R}^{C_{l}\times X_{l}\times Y_{l}}$. Subsequently, a global-local attention fusion module proposed in \cite{ming2024inversematrixvt3d} is used to merge multiscale local 3D volumes and global BEV features, resulting in multiscale fused 3D volumes $Occ_{fused}^{l} \in \mathbb{R}^{C_{l}\times X_{l}\times Y_{l}\times Z_{l}}$. These multiscale merged 3D feature volumes are further fused through upsampling and skip-connection and inputted into a UNet-like encoder-decoder to refine the features. The 3D volume logs output from the decoder $Occ_{logit}^{l}\in \mathbb{R}^{C\times X_{l}\times Y_{l}\times Z_{l}}$ are then utilized for multiscale supervision training to perform the 3D semantic occupancy prediction task. Meanwhile, drawing inspiration from DETR3D \cite{detr3d}, we include an auxiliary branch dedicated to 3D object detection. This branch involves projecting a set of trainable object queries, denoted as $Q={q_{1},q_{2},......,q_{M}}$ where $M$ is the total number of queries, onto multi-scale visual features $V$, multi-scale global BEV features $BEV_{xy}^{l}$, and multi-scale 3D volume logits $Occ_{logit}^{l}$ to aggregate features. Subsequently, these updated object queries are utilized for performing 3D object detection. The inclusion of this auxiliary branch during training introduces an additional supervision signal that effectively strengthens the capturing of small dynamic objects by enhancing both the multiscale global BEV features and the final 3D volume logits.

\subsection{Image Encoder for Surround-View Images}
The purpose of the image encoder is to capture both spatial and semantic features of the surround-view images. These features serve as the foundation for the subsequent task of predicting 3D semantic occupancy. In our approach, we first utilize a 2D backbone network (e.g. ResNet101, ResNet50) to extract visual features at multiple scales. Subsequently, these features are fused using a feature-pyramid network (FPN). The resulting visual features have resolutions that are $\frac{1}{8}$, $\frac{1}{16}$, and $\frac{1}{32}$ of the input image resolution, respectively. The deeper visual feature, with a smaller resolution, contains more semantic information and assists the model in predicting the semantic class of each voxel grid. Conversely, the relatively shallower visual feature, with larger resolutions, provides richer spatial details and better guides the model in determining whether the current voxel grid is occupied or unoccupied. Additionally, the grid mask trick, which randomly masks the grid of extracted visual features, is applied to improve the robustness of the visual features. 

\subsection{Encoder and Decoder for Merged 3D Feature Volumes}
To further enhance the quality of the merged 3D feature volumes, our approach employs an Encoder-Decoder architecture that utilizes a UNet-like network. This network refines the intermediate merged 3D feature volumes, leading to the generation of the final 3D volume logits. For the Encoder component, we implement a 3D version of ResNet18 \cite{residual}. We replace all Conv2D and BatchNorm2D operations with their 3D counterparts (Conv3D and BatchNorm3D) to refine the merged 3D feature volumes. The output of the Encoder consists of $L$ levels of multi-scale encoded 3D feature volumes, denoted as $Occ_{encoded}^{l} \in R^{C_{l}\times X_{l}\times Y_{l}\times Z_{l}}$. For the decoder component, we implement a 3D version of the Feature Pyramid Network (FPN) following a similar approach. This involves replacing all Conv2D, BatchNorm2D, and Upsample2D operations with their 3D counterparts (Conv3D, BatchNorm3D, Upsample3D). By doing so, we enable the exchange of features between the multi-scale 3D feature volumes, resulting in the generation of the final 3D volume logits. These logits are denoted as $Occ_{logit}^{l}\in \mathbb{R}^{C\times X_{l}\times Y_{l}\times Z_{l}}$, where all 3D volume logits share the same channel dimension $C$.

\subsection{3D Object Detection Auxiliary Branch}
To further enhance the model's ability to capture dynamic and partially occluded objects, we have developed an auxiliary branch for 3D object detection. This branch introduces additional 3D supervision signals during model training. Formulated as a query-based approach, we predefine a set of trainable object queries. These queries are projected onto three intermediate features sequentially (visual features $V$, multi-scale global BEV features $BEV_{xy}^{l}$, and multi-scale 3D volume logits $Occ_{logit}^{l}$) obtained from the main branch to aggregate features. The self-attention module, Visual Cross-Attention module, BEV Cross-Attention module, and 3D volume Cross-Attention module collectively constitute a query-based sampling and self-refinement block, which is iteratively stacked for $N$ layers. By utilizing the updated queries to perform and train the 3D object detection task, the information content of the intermediate features is also updated. This introduces an extra 3D supervision signal that strengthens the 3D semantic occupancy prediction task.
\subsubsection{Object Queries}
Inspired by DETR \cite{detr} and DETR3D \cite{detr3d}, we predefine a set of learnable queries $Q=\left \{ q_1,q_2,......,q_M \right \} \in R^{C} $, where $C$ represents the channel dimension of each query. From each object query $q_{i}$, we obtain the corresponding 3D point location $s_{i}\in R^{3}$ using the following method:
\begin{equation}
    s_{i} = \varphi^{sam}(q_{i})
\end{equation}
where $\varphi^{sam}$ refers to an MLP layer that generates normalized sampling locations within the range of $\left [ 0, 1\right ]$ and $s_{i}$ serve as the centre of the corresponding 3D bounding box.

\subsubsection{Self-Attention Layer}
In contrast to previous methods that employ deformable attention for self-attention in consideration of efficiency, we utilize 3D sparse convolution to enable interactions among the object queries. Specifically, we initially employ the $\varphi^{sam}$ neural network to decode the 3D point location $s_{i}$ corresponding to each object query. By decoding a set of object queries, we obtain a highly sparse point cloud $S=\left \{ s_{i}\in R^{3} \right \} _{i=1}^{M}$. Meanwhile, each query vector $Q=\left \{ q_1,q_2,......,q_M \right \} \in R^{C} $ serve as the feature vector for its corresponding 3D point. Subsequently, we apply sparse convolution to this sparse point cloud to achieve self-attention. Due to the significantly smaller number of object queries compared to the 3D volume resolution, the sparse convolution can effectively leverage the sparsity of the point cloud derived from the object queries.

\subsubsection{Visual Cross-Attention Module}
We first convert each decoded 3D point location $s_{i}$ of the query into homogeneous format $s_{i}^{\ast }$. Then, we utilize transformation matrices $T_{tran}\in R^{N\times 4\times 4}$ to project all 3D points onto multi-scale visual features $V$, as follows:
\begin{equation}
    s_{i}^{\ast }=s_{i}\oplus 1
\end{equation}
\begin{equation}
    s_{i}^{cam}=Matmul\left ( T_{tran}, s_{i}^{\ast } \right ) 
\end{equation}
where $\oplus $ refers to the concatenation operation and $Matmul$ refers to the matrix multiplication operation. During the projection of 3D points onto the visual features, we encounter points that are invalid. These include points with negative depth or coordinates outside the image resolution. Consequently, we filter out these invalid points. The remaining valid points are then divided by 8, 16, and 32, respectively, to be projected onto the corresponding scale visual features $V_{n}^{l}$. Finally, we conduct bilinear interpolation to sample visual features from each scale, culminating in a weighted sum that yields the final updated query vector. The feature sampling process can be described as follows:
\begin{equation}
    q_{i}^{updated}= \sum_{i=1}^{L} W_{i} \ast  f^{bilinear}\left ( V_{n}^{l}, s_{i}^{cam}(u, v)  \right ) 
\label{vis_ca}
\end{equation}
where $W_{i}$ is obtained as follow:
\begin{equation}
    W_{i} = MLP(q_{i} ) 
\end{equation}
The updated query $q_{i}^{updated}$ is then passed through a regression MLP to generate the $\left ( \bigtriangleup x, \bigtriangleup y, \bigtriangleup z \right )$ offset. The corresponding 3D point location is subsequently updated using the following procedure:
\begin{equation}
    \left ( \bigtriangleup x, \bigtriangleup y, \bigtriangleup z \right )=\Phi^{reg}\left ( q_{i}^{updated}  \right )
    \label{offset}
\end{equation}
\begin{equation}
    s_{i}^{updated} = s_{i} + \left ( \bigtriangleup x, \bigtriangleup y, \bigtriangleup z \right )
    \label{pts_update}
\end{equation}

\subsubsection{BEV Cross-Attention Module}
We directly utilize the updated 3D point coordinates $s_{i}^{updated}$ from the Visual Cross-Attention module as the sampling locations without the need for a projection operation. As the intermediate multi-scale global BEV features, $BEV_{xy}^{l}$ lack height information, we can perform feature sampling on BEV features using bilinear interpolation, while disregarding the $z$ dimension of $s_{i}^{updated}$. The overall sampling process is described as follows:
\begin{equation}
    q_{i}^{updated}=\sum_{i=1}^{L}W_{i}\ast f^{bilinear}\left ( BEV_{xy}^{l}, s_{i}^{updated}(x,y)  \right )  
\label{bev_ca}
\end{equation}
where $W_{i}$ is obtained as follow:
\begin{equation}
    W_{i} = MLP(q_{i}^{old} ) 
\end{equation}
and $q_{i}^{old}$ is obtained from equation \ref{vis_ca}. Subsequently, we apply the same procedure as described in equation \ref{offset} and \ref{pts_update} to update the 3D point coordinates, resulting in the new updated coordinates $s_{i}^{updated}$ associated with each updated object query $q_{i}^{updated}$.

\subsubsection{3D Volume Cross-Attention Module}
Similarly, we employ the updated 3D point coordinates $s_{i}^{updated}$ obtained from the BEV Cross-Attention Module as sampling locations for feature sampling on the intermediate 3D volume logits $Occ_{logit}^{l}$. Since $Occ_{logit}^{l}$ retains the height information, we can directly utilize bilinear interpolation in the feature sampling procedure without any modification. The complete sampling process can be described as follows:
\begin{equation}
    q_{i}^{updated}=\sum_{i=1}^{L}W_{i}\ast f^{bilinear}\left ( Occ_{logit}^{l}, s_{i}^{updated}(x,y,z)  \right )  
\end{equation}
where $W_{i}$ is obtained as follow:
\begin{equation}
    W_{i} = MLP(q_{i}^{old} ) 
\end{equation}
and $q_{i}^{old}$ is obtained from equation \ref{bev_ca}. Once again, through the repetition of the procedure outlined in equation \ref{offset} and \ref{pts_update}, we obtain the updated 3D coordinate $s_{i}^{updated}$ corresponding to each updated object query $q_{i}^{updated}$.

\begin{table*}[htpb]
  \centering
  \begin{adjustbox}{width=\textwidth}
  \begin{tabular}{c|c|c|c|cc|cccccccccccccccc}
    \toprule
    Method & Backbone & Input Modality & Params & IoU & mIoU & \rotatebox{90}{\textcolor{barrier}{$\bullet$} barrier} & \rotatebox{90}{\textcolor{bicycle}{$\bullet$} bicycle} & \rotatebox{90}{\textcolor{bus}{$\bullet$} bus} & \rotatebox{90}{\textcolor{car}{$\bullet$} car} & \rotatebox{90}{\textcolor{construction}{$\bullet$} const. veh.} & \rotatebox{90}{\textcolor{motorcycle}{$\bullet$} motorcycle} & \rotatebox{90}{\textcolor{pedestrian}{$\bullet$} pedestrian} & \rotatebox{90}{\textcolor{cone}{$\bullet$} traffic cone} & \rotatebox{90}{\textcolor{trailer}{$\bullet$} trailer} & \rotatebox{90}{\textcolor{truck}{$\bullet$} truck} & \rotatebox{90}{\textcolor{driveable}{$\bullet$} drive. surf.} & \rotatebox{90}{\textcolor{flat}{$\bullet$} other flat} & \rotatebox{90}{\textcolor{sidewalk}{$\bullet$} sidewalk} & \rotatebox{90}{\textcolor{terrain}{$\bullet$} terrain} & \rotatebox{90}{\textcolor{manmade}{$\bullet$} manmade} & \rotatebox{90}{\textcolor{vegetation}{$\bullet$} vegetation} \\
     \midrule 
    MonoScene\cite{monoscene} & ResNet101-DCN & C & - & 23.96 & 7.31 & 4.03 & 0.35 & 8.00 & 8.04 & 2.90 & 0.28 & 1.16 & 0.67 & 4.01 & 4.35 & 27.72 & 5.20 & 15.13 & 11.29 & 9.03 & 14.86\\
    Atlas* \cite{atlas} & - & C & - & 28.66 & 15.00 & 10.64 & 5.68 & 19.66 & 24.94 & 8.90 & 8.84 & 6.47 & 3.28 & 10.42 & 16.21 & 34.86 & 15.46 & 21.89 & 20.95 & 11.21 & 20.54 \\
    BEVFormer*\cite{bevformer} & ResNet101-DCN & C & 59M & 30.50 & 16.75 & 14.22 & 6.58 & 23.46 & 28.28 & 8.66 & 10.77 & 6.64 & 4.05 & 11.20 & 17.78 & 37.28 & 18.00 & 22.88 & 22.17 & 13.80 & 22.21 \\
    TPVFormer\cite{tpvformer} & ResNet101-DCN & C & 69M  & 11.51 & 11.66 & 16.14 & 7.17 & 22.63 & 17.13 & 8.83 & 11.39 & 10.46 & 8.23 & 9.43 & 17.02 & 8.07 & 13.64 & 13.85 & 10.34 & 4.90 & 7.37 \\
    TPVFormer* & ResNet101-DCN & C & 69M & 30.86 & 17.10 & 15.96 & 5.31 & 23.86 & 27.32 & 9.79 & 8.74 & 7.09 & 5.20 & 10.97 & 19.22 & 38.87 & 21.25 & 24.26 & 23.15 & 11.73 & 20.81 \\
    C-CONet*\cite{openoccupancy} & ResNet101 & C & 118M & 26.10 & 18.40 & 18.60 & 10.00 & 26.40 & 27.40 & 8.60 & 15.70 & 13.30 & 9.70 & 10.90 & 20.20 & 33.00 & 20.70 & 21.40 & 21.80 & 14.70 & 21.30 \\
    InverseMatrixVT3D*\cite{ming2024inversematrixvt3d} & ResNet101-DCN & C & 67M & 
    30.03 & 18.88 & 18.39 & 12.46 & 26.30 & 29.11 & 11.00 & 15.74 & 14.78 & 11.38 & 13.31 & 21.61 & 36.30 & 19.97 & 21.26 & 20.43 & 11.49 & 18.47 \\
    OccFormer*\cite{occformer} & ResNet101-DCN & C & 169M & 31.39 & 19.03 & 18.65 & 10.41 & 23.92 & 30.29 & 10.31 & 14.19 & 13.59 & 10.13 & 12.49 & 20.77 & 38.78 & 19.79 & 24.19 & 22.21 & 13.48 & 21.35 \\
    FB-Occ*\cite{fbocc} & ResNet101 & C & - & 31.50 & 19.60 & 20.60 & 11.30 & 26.90 & 29.80 & 10.40 & 13.60 & 13.70 & 11.40 & 11.50 & 20.60 & 38.20 & 21.50 & 24.60 & 22.70 & 14.80 & 21.60 \\
    RenderOcc*\cite{renderocc} & ResNet101 & C & 122M & 29.20 & 19.00 & 19.70 & 11.20 & 28.10 & 28.20 & 9.80 & 14.70 & 11.80 & 11.90 & 13.10 & 20.10 & 33.20 & 21.30 & 22.60 & 22.30 & 15.30 & 20.90 \\
    GaussianFormer*\cite{gaussianformer} & ResNet101-DCN & C & - & 29.83 & 19.10 & 19.52 & 11.26 & 26.11 & 29.78 & 10.47 & 13.83 & 12.58 & 8.67 & 12.74 & 21.57 & 39.63 & 23.28 & 24.46 & 22.99 & 9.59 & 19.12 \\
    Co-Occ*\cite{co-occ} & ResNet101 & C & 218M & 30.00 & 20.30 & \textbf{22.50} & 11.20 & \textbf{28.60} & 29.50 & 9.90 & 15.80 & 13.50 & 8.70 & 13.60 & 22.20 & 34.90 & 23.10 & 24.20 & \textbf{24.10} & \textbf{18.00} & \textbf{24.80} \\
    GaussianFormer2-256*\cite{gaussianformer2} & ResNet101-DCN & C & - & 31.14 & 20.36 & 19.93 & 12.99 & 28.15 & 30.82 & 10.97 & 16.54 & 13.23 & 10.56 & 13.39 & 22.20 & \textbf{39.71} & 23.65 & \textbf{25.43} & 23.68 & 12.96 & 21.51 \\
    SurroundOcc*\cite{surroundocc} & ResNet101-DCN & C & 180M & 31.49 & 20.30 & 20.59 & 11.68 & 28.06 & 30.86 & 10.70 & 15.14 & 14.09 & 12.06 & \textbf{14.38} & 22.26 & 37.29 & \textbf{23.70} & 24.49 & 22.77 & 14.89 & 21.86 \\
    \hline
    Inverse++* (ours) & ResNet101-DCN & C & 137M & \textbf{31.73} & \textbf{20.91} & 20.90 & \textbf{13.27} & 28.40 & \textbf{31.37} & \textbf{11.90} & \textbf{17.76} & \textbf{15.39} & \textbf{13.49} & 13.32 & \textbf{23.19} & 39.37 & 22.85 & 25.27 & 23.68 & 13.43 & 20.98 \\
    \hline\hline
    \rowcolor{gray!30} LMSCNet*\cite{lmscnet} & - & L &  - & 36.60 & 14.90 & 13.10 & 4.50 & 14.70 & 22.10 & 12.60 & 4.20 & 7.20 & 7.10 & 12.20 & 11.50 & 26.30 & 14.30 & 21.10 & 15.20 & 18.50 & 34.20 \\
    \rowcolor{gray!30} L-CONet*\cite{openoccupancy} & - & L & - & 39.40 & 17.70 & 19.20 & 4.00 & 15.10 & 26.90 & 6.20 & 3.80 & 6.80 & 6.00 & 14.10 & 13.10 & 39.70 & 19.10 & 24.00 & 23.90 & 25.10 & 35.70 \\
    \rowcolor{gray!30} OccFusion (C+R)*\cite{occfusion} & R101-DCN+VoxelNet & C+R & - & 32.90 & 20.73 & 20.46 & 13.98 & 27.99 & 31.52 & 13.68 & 18.45 & 15.79 & 13.05 & 13.94 & 23.84 & 37.85 & 19.60 & 22.41 & 21.20 & 16.16 & 21.81 \\
    \bottomrule
  \end{tabular}
  \end{adjustbox}
  \caption{\textbf{3D semantic occupancy prediction results on SurroundOcc-nuScenes validation set}. Our approach outperforms other existing methods with the same input modality. For readers' reference, the bottom of the table presents results from three additional methods using different input modalities. * means method is trained with dense occupancy labels from SurroundOcc \cite{surroundocc}. Notion of modality: Camera (C), Lidar (L), Radar (R).}
  \label{occ}
\end{table*}

\section{Experimental Results}\label{simulation}
\subsection{Implementation Details}
The Inverse++ model incorporates ResNet101-DCN \cite{residual} and FPN \cite{fpn} for its image encoder. The features from stages 1, 2, and 3 of ResNet101-DCN are passed to FPN \cite{fpn}, generating three levels of multi-scale visual features. The query-based sampling and self-refinement block in the auxiliary branch, comprised of a Self-Attention layer, a Visual Cross-Attention Module, a BEV Cross-Attention Module, and a 3D volume Cross-Attention Module, is iteratively stacked six times. The AdamW optimizer is utilized for optimization, with an initial learning rate of 2e-4 and weight decay of 0.01. The learning rate is decayed using a multi-step scheduler. For data augmentation, random resize, rotation, and flip operations are implemented in the image space, following established practices for BEV-based 3D object detection \cite{bevdet,bevdepth,bevstereo,bevformer} and the compared methods \cite{tpvformer, surroundocc, occformer, monoscene}. The predicted occupancy has a resolution of $200 \times 200 \times 16$ for full-scale evaluation. Training of the model is conducted on three A40 GPUs with 48GB of memory, spanning a duration of 5 days.

\subsection{Loss Function}
To train the model with both main and auxiliary branches, we employ focal loss \cite{focal}, Lovasz-softmax loss \cite{lovasz}, and scene-class affinity loss \cite{monoscene} to address the significant sparsity of free space in the 3D semantic occupancy prediction task. For the auxiliary task of 3D object detection, we utilize focal loss for class label classification and L1 loss for bounding box parameter regression, following the methodology of DETR3D. The final loss is composed of:
\begin{equation}
    Occ_{Loss} = \sum_{l=1}^{L+1} \frac{1}{2^{l} } \times (L_{focal}^{l} + L_{lovasz}^{l} + L_{scal\_geo}^{l} + L_{scal\_sem}^{l})
\end{equation}
\begin{equation}
    Det_{Loss} = \sum_{n=1}^{N}\sum_{j=1}^{3}L_{focal}^{j} + L_{L1}^{j}
\end{equation}
\begin{equation}
    Loss=Det_{Loss}+\lambda  Occ_{Loss} 
\end{equation}
where $\lambda$ balances the loss weight between main and auxiliary branches. In practice, the parameter values are set to $\lambda=2$ and $L=3$. The training phase involves supervising the output of the Visual Cross-Attention Module, BEV Cross-Attention Module, and 3D volume Cross-Attention Module. Moreover, the query-based sampling and self-refinement block will be stacked 6 times as $N=6$.

\subsection{Dataset}
The public nuScenes dataset \cite{nuscenes}, specifically designed for autonomous driving purposes, serves as the primary data source for our experiments. To perform the 3D semantic occupancy prediction task, we utilize dense labels obtained from SurroundOcc \cite{surroundocc}. 
Since the test set lacks semantic labels, we train our model on the training set and evaluate its performance using the validation set. For 3D semantic occupancy prediction using annotations from SurroundOcc, we set the range of the X and Y axes to [-50, 50] meters and the Z axis to [-5, 3] meters under lidar coordinates. 
The input images have a resolution of $1600 \times 900$ pixels, while the final output of the semantic occupancy prediction branch is represented with a resolution of $200 \times 200 \times 16$. The annotations from SurroundOcc contain a total of 17 semantic classes with label 0 refer to free voxel. 
On the other hand, the auxiliary 3D object detection branch yields 9-dimensional parameters (x, y, z, l, h, w, yaw, vx, vy) representing the centre, length, width, height, yaw angle, and velocity along the x and y axes of the bounding box. Additionally, following the methodology proposed in \cite{occfusion}, we conduct an in-depth analysis of our model's performance in challenging scenarios, specifically rainy and nighttime conditions. This evaluation is carried out using the annotation file provided by \cite{occfusion}.

\subsection{Performance Evaluate Metrics}
To assess the performance of various state-of-the-art (SOTA) algorithms and compare them with our approach in the 3D semantic occupancy prediction task, we utilize the intersection over union (IoU) to evaluate each semantic class. Moreover, we employ the mean IoU overall semantic classes (mIoU) as a comprehensive evaluation metric:
\begin{equation}
    IoU=\frac{TP}{TP+FP+FN} 
\end{equation}
and
\begin{equation}
    mIoU=\frac{1}{Cls}\sum_{i=1}^{Cls}  \frac{TP_{i}}{TP_{i}+FP_{i}+FN_{i}} 
\end{equation}
where $TP$, $FP$, and $FN$ represent the counts of true positives, false positives, and false negatives in our predictions, respectively, while $Cls$ denotes the total class number.

\subsection{Model Performance Analysis}
To evaluate the performance of our proposed model, Inverse++, we compared it with other state-of-the-art algorithms and presented the results in Table \ref{occ}. 
In Table \ref{occ}, our model exhibited highly competitive performance, outperforming previous vision-centric state-of-the-art methods and ranking first on the benchmark according to the IoU and mIoU evaluation metrics. Our method even outperforms OccFusion(C+R), a multi-modality fusion approach, under the mIoU evaluation metric. Notably, our model incorporates a 3D object detection auxiliary branch that introduces additional supervision signals on intermediate features, allowing it to excel in capturing small dynamic objects on the road, such as bicycles, motorcycles, and pedestrians, who are all vulnerable road users. Additionally, our model also achieves outstanding performance in detecting general dynamic objects on the road, including buses, cars, construction vehicles, trailers, and trucks. It's worth mentioning that despite having only 135M trainable parameters, substantially fewer than SurroundOcc and other similar performance methods, our model still outperforms them.


\subsection{Challenging Scenarios Performance Analysis}
\begin{table*}[htpb]
  \centering
  \begin{adjustbox}{width=\textwidth}
  \begin{tabular}{c|c|c|cc|cccccccccccccccc}
    \toprule
    Method & Backbone & Input Modality & IoU & mIoU & \rotatebox{90}{\textcolor{barrier}{$\bullet$} barrier} & \rotatebox{90}{\textcolor{bicycle}{$\bullet$} bicycle} & \rotatebox{90}{\textcolor{bus}{$\bullet$} bus} & \rotatebox{90}{\textcolor{car}{$\bullet$} car} & \rotatebox{90}{\textcolor{construction}{$\bullet$} const. veh.} & \rotatebox{90}{\textcolor{motorcycle}{$\bullet$} motorcycle} & \rotatebox{90}{\textcolor{pedestrian}{$\bullet$} pedestrian} & \rotatebox{90}{\textcolor{cone}{$\bullet$} traffic cone} & \rotatebox{90}{\textcolor{trailer}{$\bullet$} trailer} & \rotatebox{90}{\textcolor{truck}{$\bullet$} truck} & \rotatebox{90}{\textcolor{driveable}{$\bullet$} drive. surf.} & \rotatebox{90}{\textcolor{flat}{$\bullet$} other flat} & \rotatebox{90}{\textcolor{sidewalk}{$\bullet$} sidewalk} & \rotatebox{90}{\textcolor{terrain}{$\bullet$} terrain} & \rotatebox{90}{\textcolor{manmade}{$\bullet$} manmade} & \rotatebox{90}{\textcolor{vegetation}{$\bullet$} vegetation} \\
     \midrule 
    InverseMatrixVT3D\cite{ming2024inversematrixvt3d} & R101-DCN & C & 29.72 & 18.99 & 18.55 & 14.29 & 22.28 & 30.02 & 10.19 & 15.20 & 10.03 & 9.71 & 13.28 & 20.98 & 37.18 & 23.47 & 27.74 & 17.46 & 10.36 & 23.13 \\
    GaussianFormer\cite{gaussianformer} & R101-DCN & C & 27.37 & 16.96 & 18.16 & 9.58 & 21.09 & 26.83 & 8.04 & 10.13 & 7.80 & 5.84 & 12.66 & 18.24 & 35.53 & 18.51 & 27.79 & 19.23 & 11.04 & 20.85 \\
    Co-Occ \cite{co-occ} & R101 & C & 28.90 & 19.70 & 22.10 & \textbf{17.60} & 26.30 & 30.80 & 10.90 & 9.90 & 8.20 & 9.70 & 11.40 & 19.30 & 39.00 & 22.20 & \textbf{32.60} & \textbf{23.00} & 11.50 & 21.30 \\
    GaussianFormer2-256\cite{gaussianformer2} & R101-DCN & C & 31.14 & 20.36 & 19.84 & 13.52 & \textbf{26.89} & 31.65 & 10.82 & 15.16 & 9.04 & 8.41 & 13.72 & 21.84 & \textbf{40.51} & 24.57 & 32.21 & 20.65 & 12.64 & 24.33 \\
    SurroundOcc \cite{surroundocc} & R101-DCN & C & 30.57 & 19.85 & 21.40 & 12.75 & 25.49 & 31.31 & 11.39 & 12.65 & 8.94 & 9.48 & \textbf{14.51} & 21.52 & 35.34 & \textbf{25.32} & 29.89 & 18.37 & \textbf{14.44} & \textbf{24.78} \\
    \hline
    Inverse++ & R101-DCN & C & \textbf{31.32} & \textbf{20.66} & \textbf{22.52} & 13.79 & 25.49 & \textbf{31.80} & \textbf{11.70} & \textbf{16.72} & \textbf{11.14} & \textbf{10.12} & 12.29 & \textbf{22.25} & 38.78 & 23.93 & 31.62 & 21.14 & 12.65 & 24.61 \\
    \bottomrule
  \end{tabular}
  \end{adjustbox}
  \caption{\textbf{3D semantic occupancy prediction results on SurroundOcc-nuScenes validation rainy scenario subset}. All methods are trained with dense occupancy labels from SurroundOcc \cite{surroundocc}. Notion of modality: Camera (C).}
  \label{occ_rainy}
\end{table*}

\begin{table*}[htpb]
  \centering
  \begin{adjustbox}{width=\textwidth}
  \begin{tabular}{c|c|c|cc|cccccccccccccccc}
    \toprule
    Method & Backbone & Input Modality & IoU & mIoU & \rotatebox{90}{\textcolor{barrier}{$\bullet$} barrier} & \rotatebox{90}{\textcolor{bicycle}{$\bullet$} bicycle} & \rotatebox{90}{\textcolor{bus}{$\bullet$} bus} & \rotatebox{90}{\textcolor{car}{$\bullet$} car} & \rotatebox{90}{\textcolor{construction}{$\bullet$} const. veh.} & \rotatebox{90}{\textcolor{motorcycle}{$\bullet$} motorcycle} & \rotatebox{90}{\textcolor{pedestrian}{$\bullet$} pedestrian} & \rotatebox{90}{\textcolor{cone}{$\bullet$} traffic cone} & \rotatebox{90}{\textcolor{trailer}{$\bullet$} trailer} & \rotatebox{90}{\textcolor{truck}{$\bullet$} truck} & \rotatebox{90}{\textcolor{driveable}{$\bullet$} drive. surf.} & \rotatebox{90}{\textcolor{flat}{$\bullet$} other flat} & \rotatebox{90}{\textcolor{sidewalk}{$\bullet$} sidewalk} & \rotatebox{90}{\textcolor{terrain}{$\bullet$} terrain} & \rotatebox{90}{\textcolor{manmade}{$\bullet$} manmade} & \rotatebox{90}{\textcolor{vegetation}{$\bullet$} vegetation} \\
     \midrule 
    InverseMatrixVT3D\cite{ming2024inversematrixvt3d} & R101-DCN & C & 22.48 & 9.99 & 10.40 & 12.03 & 0.00 & 29.94 & 0.00 & 9.92 & 4.88 & \textbf{0.91} & 0.00 & 17.79 & 29.10 & 2.37 & 10.80 & 9.40 & 8.68 & 13.57 \\
    GaussianFormer\cite{gaussianformer} & R101-DCN & C & 20.30 & 9.07 & 6.11 & 8.70 & 0.00 & 25.75 & 0.00 & 10.44 & 2.85 & 0.55 & 0.00 & 17.26 & 30.65 & \textbf{2.95} & 12.53 & 9.94 & 6.65 & 10.71 \\
    Co-Occ \cite{co-occ} & R101 & C & 18.90 & 9.40 & 4.50 & 9.30 & 0.00 & 29.50 & 0.00 & 8.40 & 3.50 & 0.00 & 0.00 & 15.10 & 29.40 & 0.60 & 12.40 & 11.50 & \textbf{10.70} & 15.50 \\
    GaussianFormer2-256\cite{gaussianformer2} & R101-DCN & C & 21.19 & 10.14 & 5.25 & 9.29 & 0.00 & 29.33 & 0.00 & \textbf{13.65} & 5.80 & 0.90 & 0.00 & 20.22 & 31.80 & 1.94 & \textbf{14.83} & 10.48 & 5.96 & 12.72 \\
    SurroundOcc \cite{surroundocc} & R101-DCN & C & \textbf{24.38} & 10.80 & \textbf{10.55} & \textbf{14.60} & 0.00 & 31.05 & 0.00 & 8.26 & 5.37 & 0.58 & 0.00 & 18.75 & 30.72 & 2.74 & 12.39 & \textbf{11.53} & 10.52 & \textbf{15.77} \\
    \hline
    Inverse++ & R101-DCN & C & 23.70 & \textbf{10.93} & 8.87 & 10.19 & 0.00 & \textbf{32.62} & 0.00 & 11.77 & \textbf{7.46} & 0.72 & 0.00 & \textbf{22.20} & \textbf{32.95} & 2.15 & 13.01 & 9.79 & 8.61 & 14.48 \\
    \bottomrule
  \end{tabular}
  \end{adjustbox}
  \caption{\textbf{3D semantic occupancy prediction results on SurroundOcc-nuScenes validation night scenario subset}. All methods are trained with dense occupancy labels from SurroundOcc \cite{surroundocc}. Notion of modality: Camera (C).}
  \label{occ_night}
\end{table*}

\begin{figure*}[htbp]
     \centering
     \begin{subfigure}[]{0.32\textwidth}
         \centering
         \includegraphics[width=\textwidth]{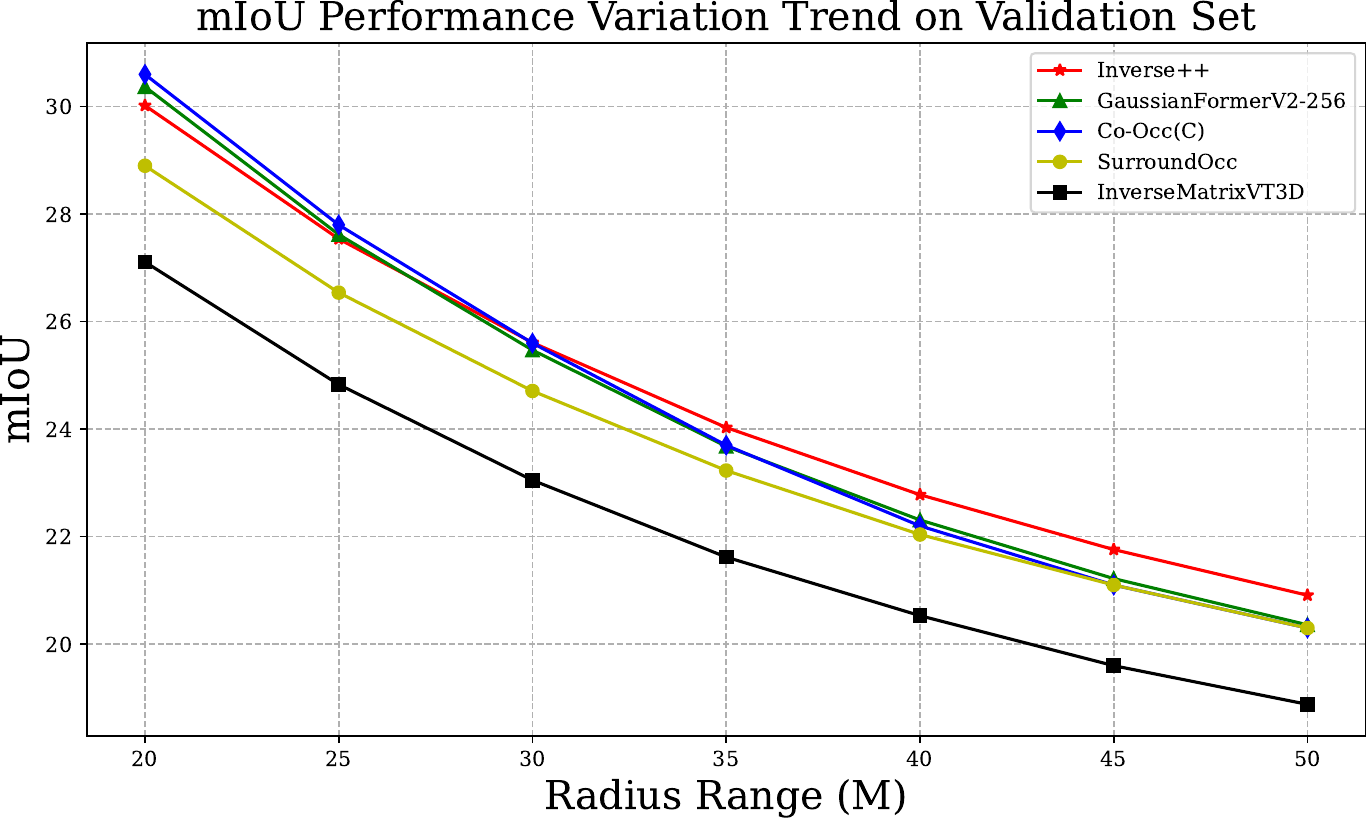}
         \caption{}
         \label{mIoU_whole_trend}
     \end{subfigure}
     \begin{subfigure}[]{0.33\textwidth}
         \centering
         \includegraphics[width=\textwidth]{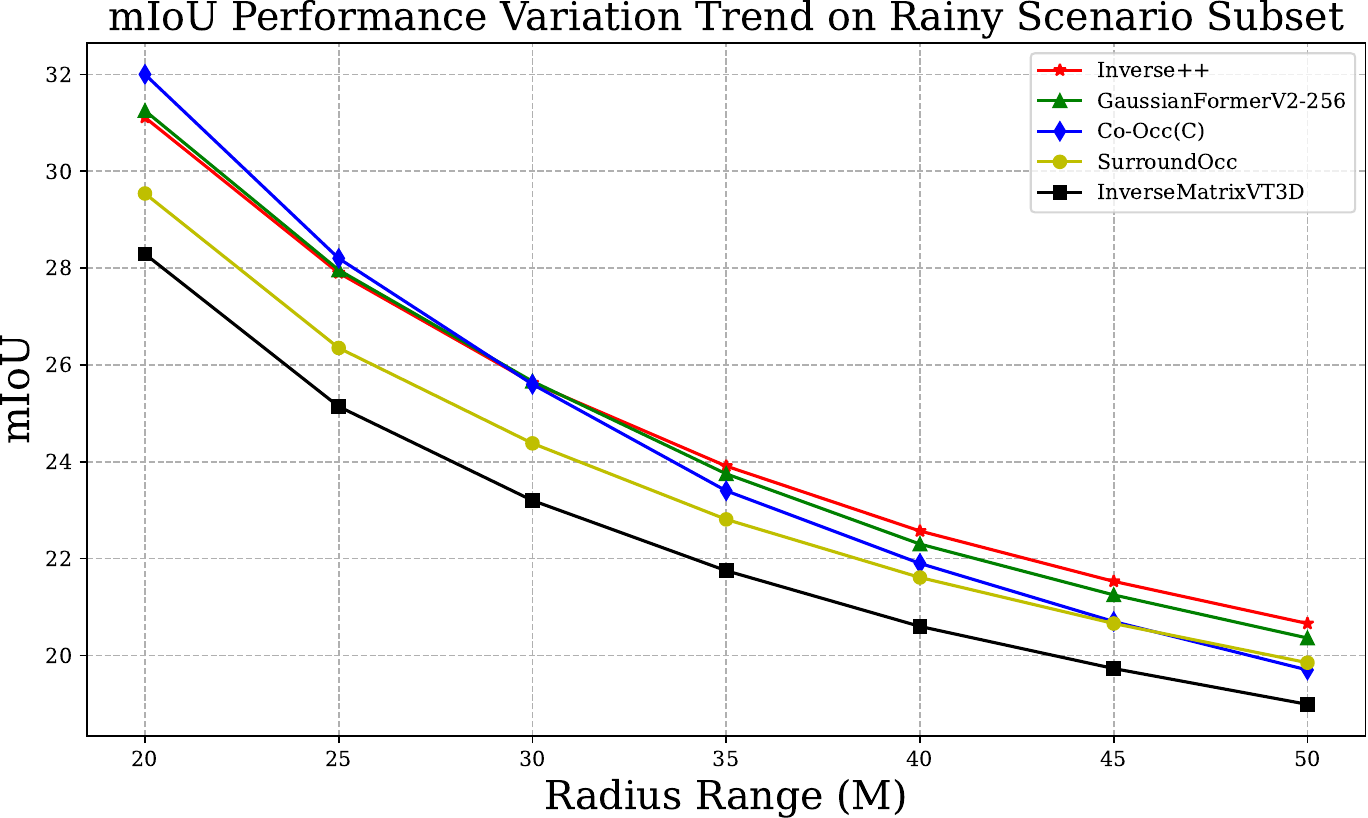}
         \caption{}
         \label{mIoU_rainy_trend}
     \end{subfigure}
     \begin{subfigure}[]{0.33\textwidth}
         \centering
         \includegraphics[width=\textwidth]{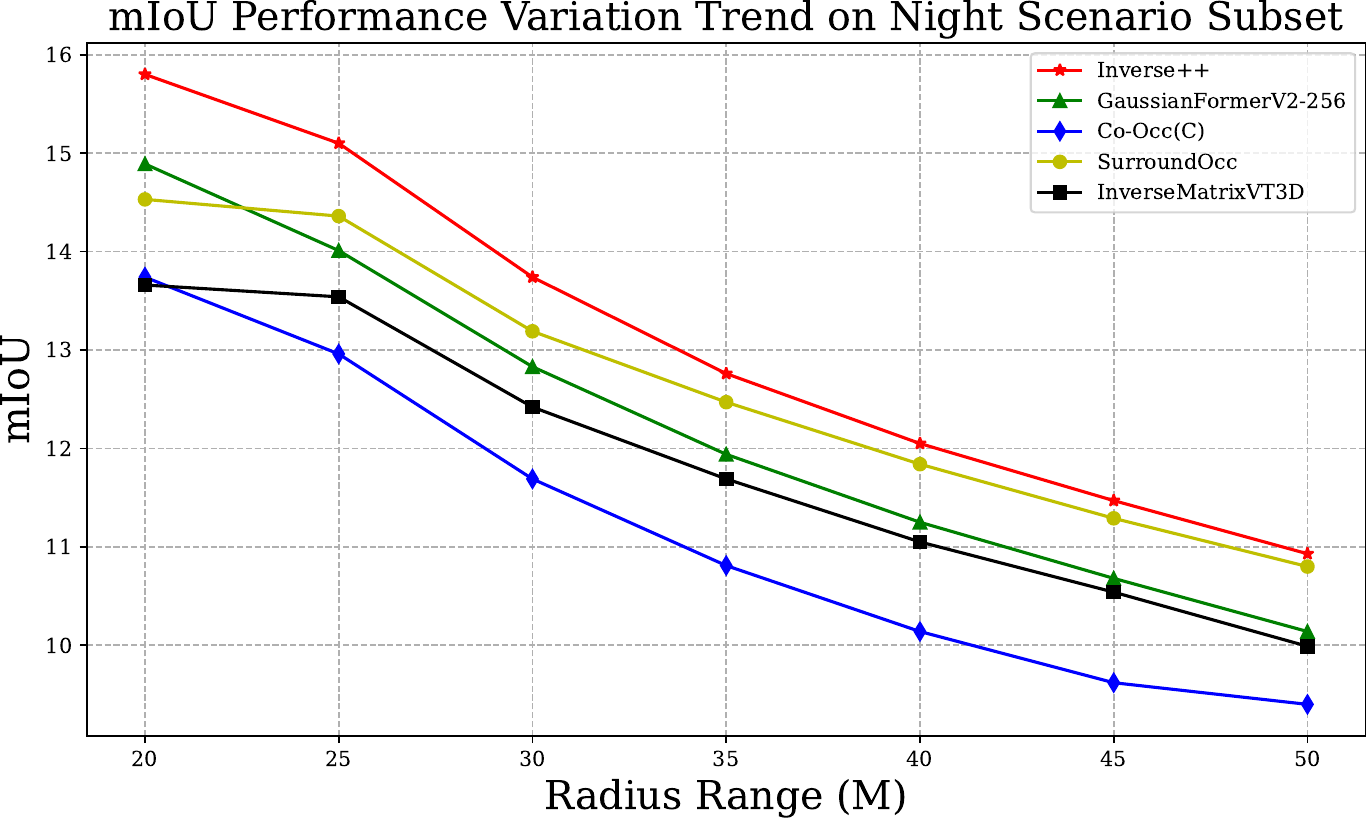}
         \caption{}
         \label{mIoU_night_trend}
     \end{subfigure}
     \begin{subfigure}[]{0.32\textwidth}
         \centering
         \includegraphics[width=\textwidth]{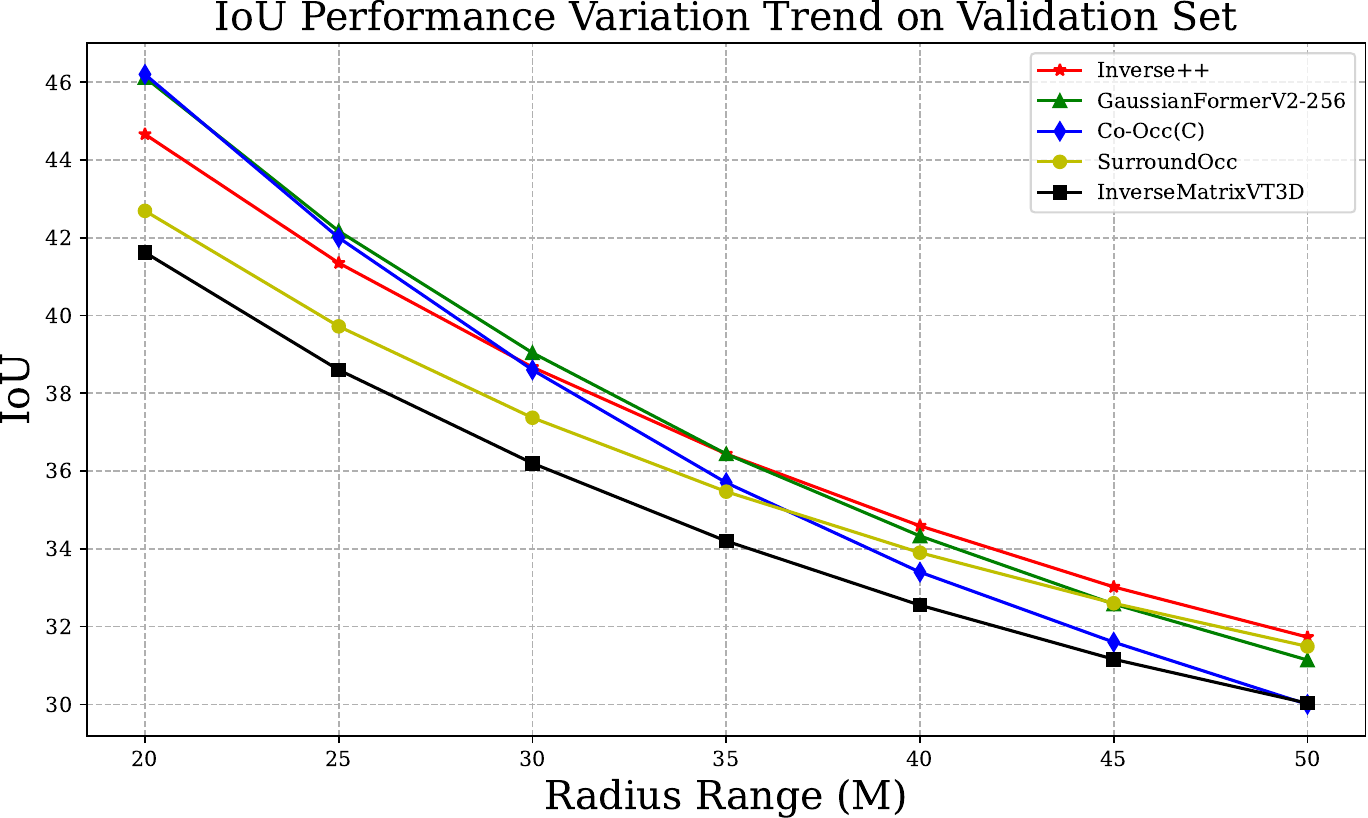}
         \caption{}
         \label{IoU_whole_trend}
     \end{subfigure}
     \begin{subfigure}[]{0.33\textwidth}
         \centering
         \includegraphics[width=\textwidth]{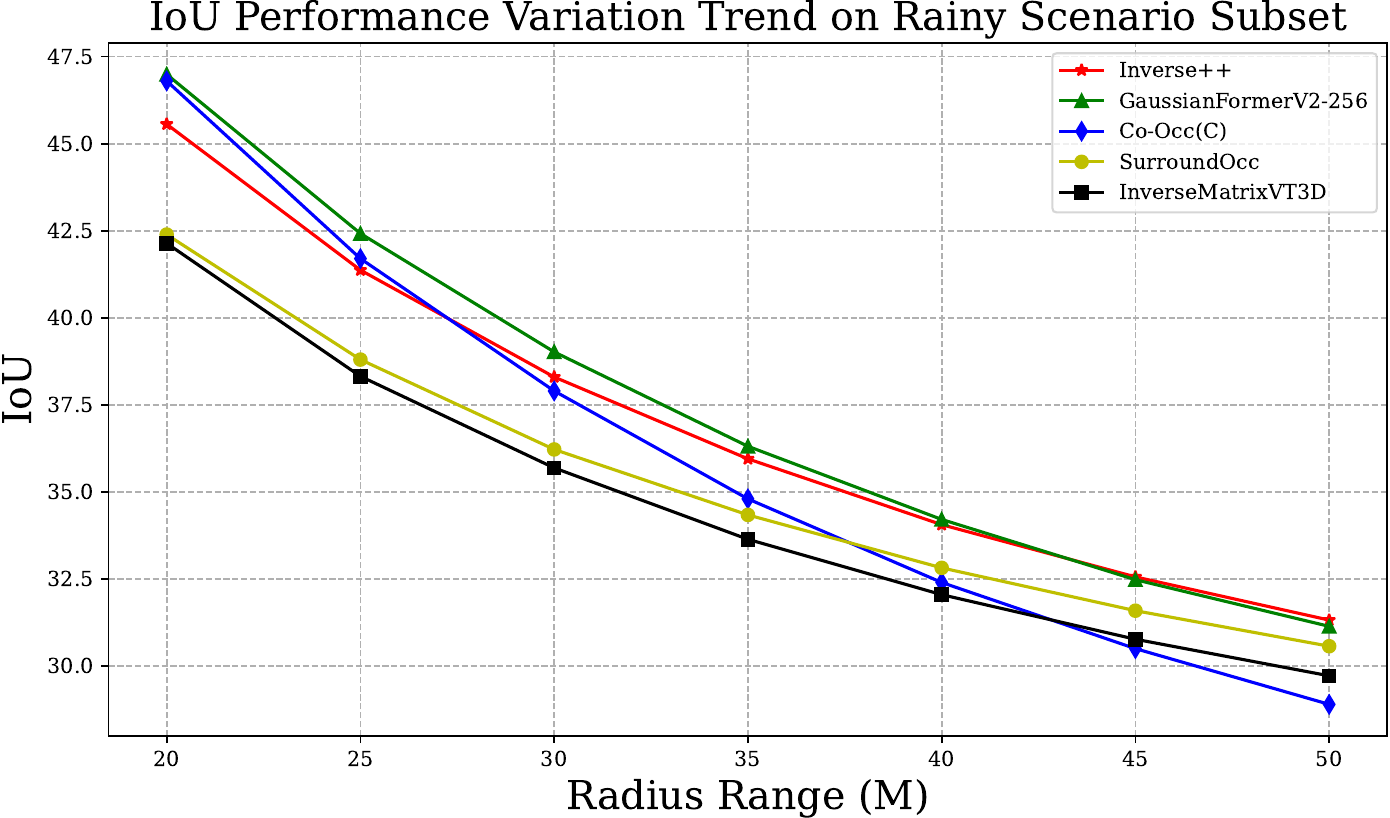}
         \caption{}
         \label{IoU_rainy_trend}
     \end{subfigure}
     \begin{subfigure}[]{0.33\textwidth}
         \centering
         \includegraphics[width=\textwidth]{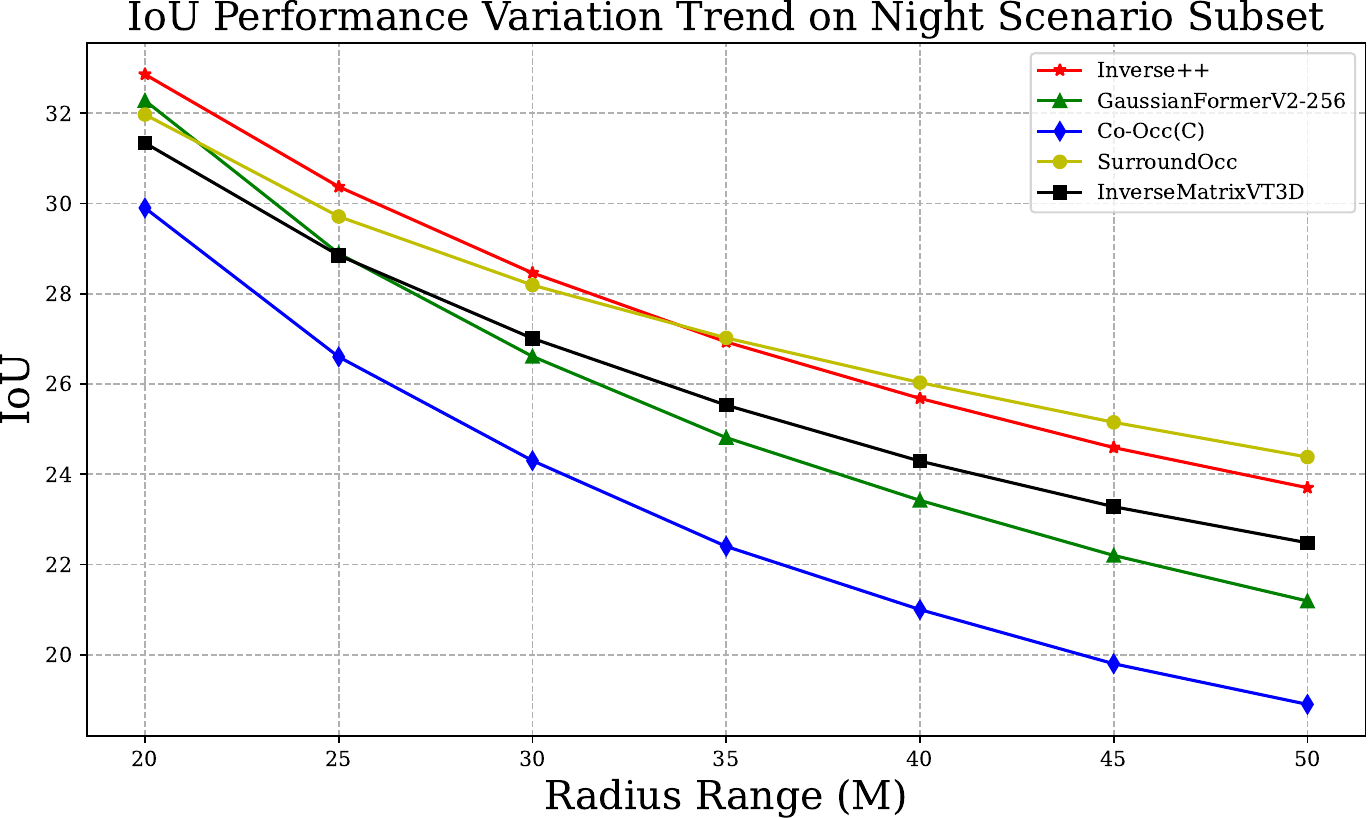}
         \caption{}
         \label{IoU_night_trend}
     \end{subfigure}
        \caption{\small Performance variation trend for 3D semantic occupancy prediction task. (a) mIoU performance variation trend on the whole SurroundOcc-nuScenes validation set, (b) mIoU performance variation trend on the SurroundOcc-nuScenes validation rainy scenario subset, and (c) mIoU performance variation on the SurroundOcc-nuScenes validation night scenario subset. (d) IoU performance variation on the whole SurroundOcc-nuScenes validation set, (e) IoU performance variation on the SurroundOcc-nuScenes validation rainy scenario subset, and (f) IoU performance variation on the SurroundOcc-nuScenes validation night scenario subset. \textbf{Better viewed when zoomed in.}}
        \label{perform_var}
\end{figure*}

To comprehensively assess the capability and robustness of our model in challenging scenarios like rain and nighttime, we adopt the methodology and utilize the annotation files proposed in \cite{occfusion} to evaluate the performance of our model. We compare our model with other state-of-the-art methods in terms of its performance in rainy and nighttime scenarios. The results for the model's performance in rainy and nighttime scenarios are presented in Table \ref{occ_rainy} and Table \ref{occ_night}, respectively.

For the rainy scenarios, all algorithms have experienced varying degrees of performance degradation. Despite this, our algorithm performs best among all the degradation algorithms, demonstrating the best robustness of our algorithm in those SOTA methods under rainy scenarios.

In nighttime scenarios, all state-of-the-art algorithms suffer from significant performance degradation due to the camera sensor's sensitivity to ambient lighting conditions. However, our algorithm experiences the least amount of performance degradation compared to all the other SOTA algorithms.

\subsection{Performance Analysis On Varying Distance}
The additional 3D supervision signal introduced by the auxiliary branch enhanced overall performance and alleviated algorithm performance degradation over distance. In this study, we examine our algorithm's performance along with other SOTA algorithms under different perception ranges in different scenarios. Each algorithm is evaluated at perception range at $R=[20m,25m,30m,35m,40m,45m,50m]$. 

Performance variation trend on the whole SurroundOcc-nuScenes validation set is demonstrated in Figure \ref{mIoU_whole_trend} and Figure \ref{IoU_whole_trend}. Our algorithm only achieved competitive performance when the perception range was short. However, as the perception range increased, all SOTA algorithms except ours experienced relatively fast performance degradation. Figure \ref{mIoU_rainy_trend} and Figure \ref{IoU_rainy_trend} depict the mIoU and IoU performance variation trend in the rainy scenario. The overall variation trend is similar to the variation trend on the whole validation set, and we observed a fast performance decay of Co-Occ(C) under this scenario as the perception range increased. In the nighttime scenario, the mIoU and IoU performance variation trend is shown in Figure \ref{mIoU_night_trend} and Figure \ref{IoU_night_trend}. Except for SurroundOcc and our algorithm, all other algorithms experienced severe performance degradation as the perception range increased.

\subsection{Challenging Scenes Qualitative Analysis}
\begin{figure*}[htbp]
\centering
\begin{subfigure}[]{\textwidth}
\includegraphics[width=\textwidth]{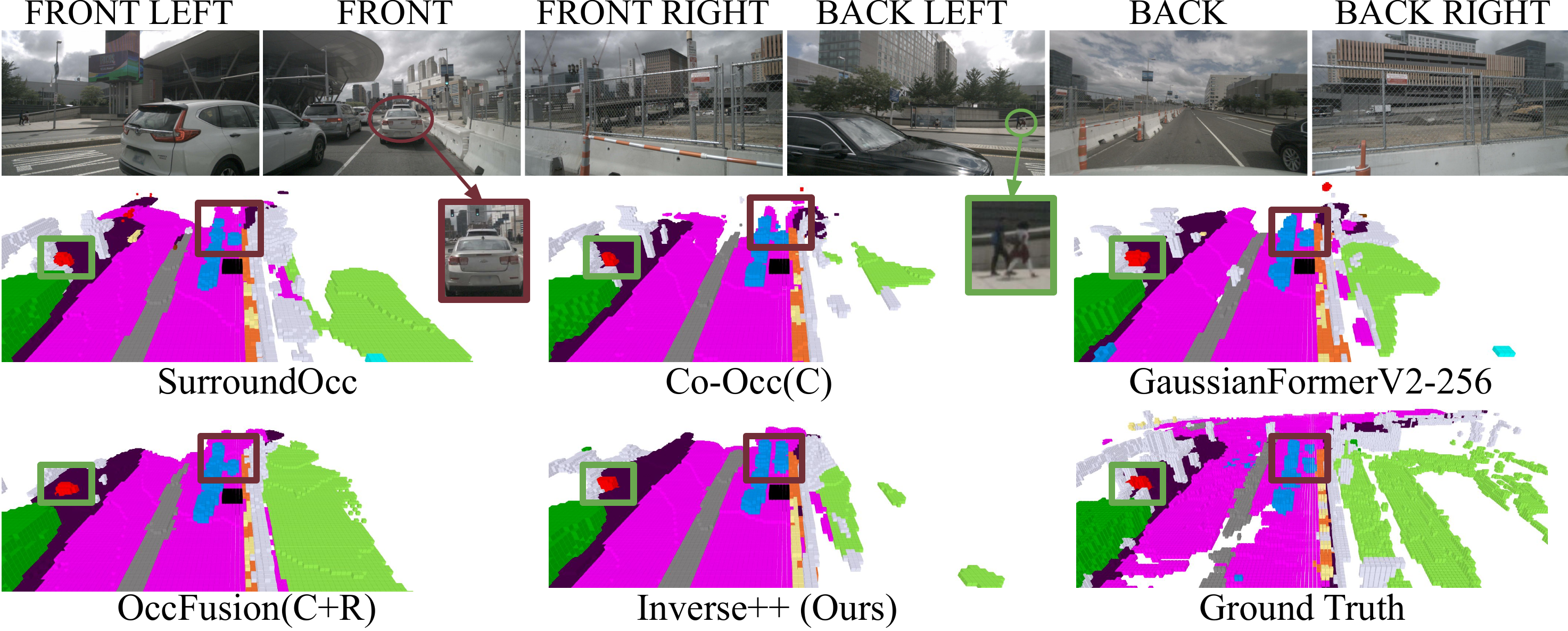}\\
\includegraphics[width=\textwidth]{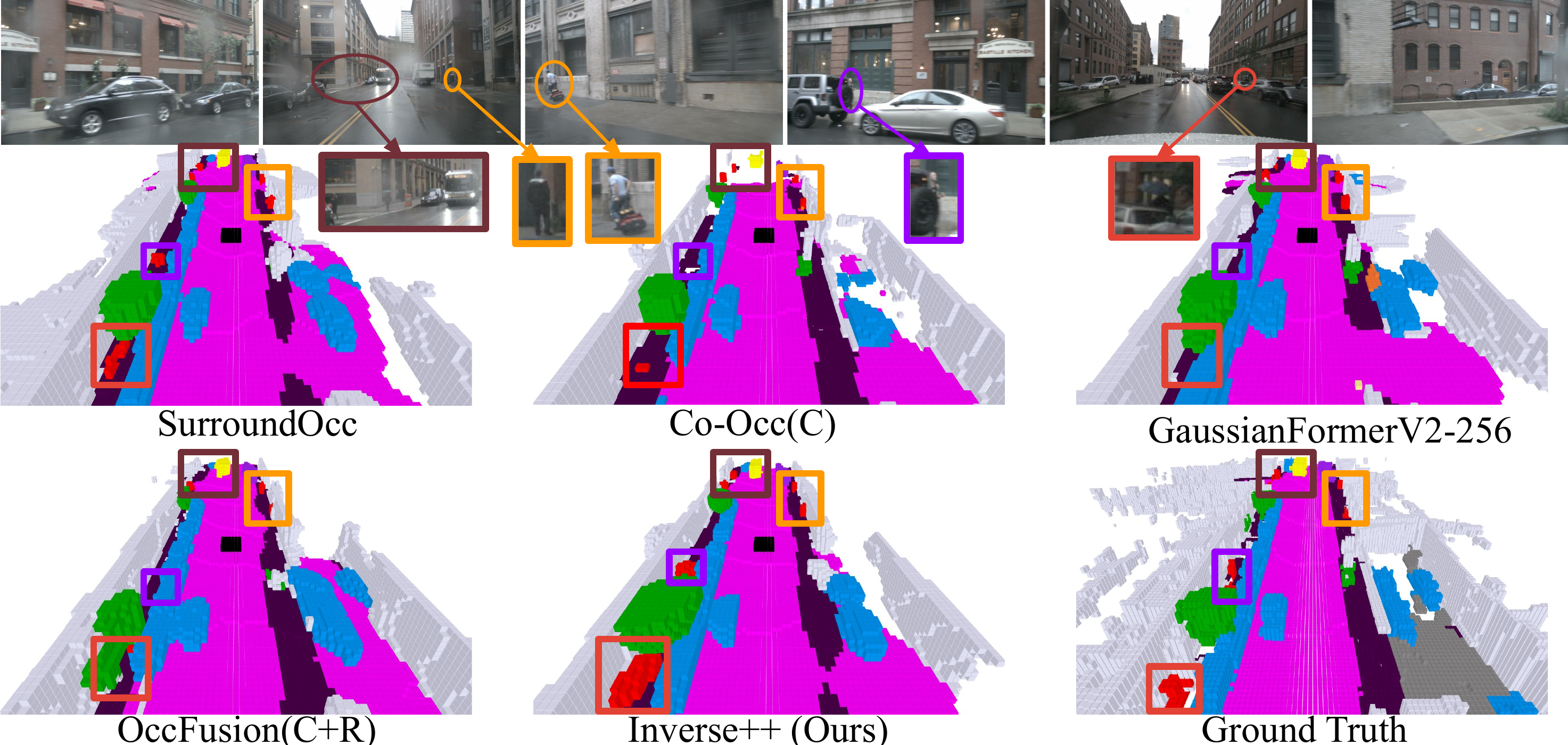}\\
\includegraphics[width=\textwidth]{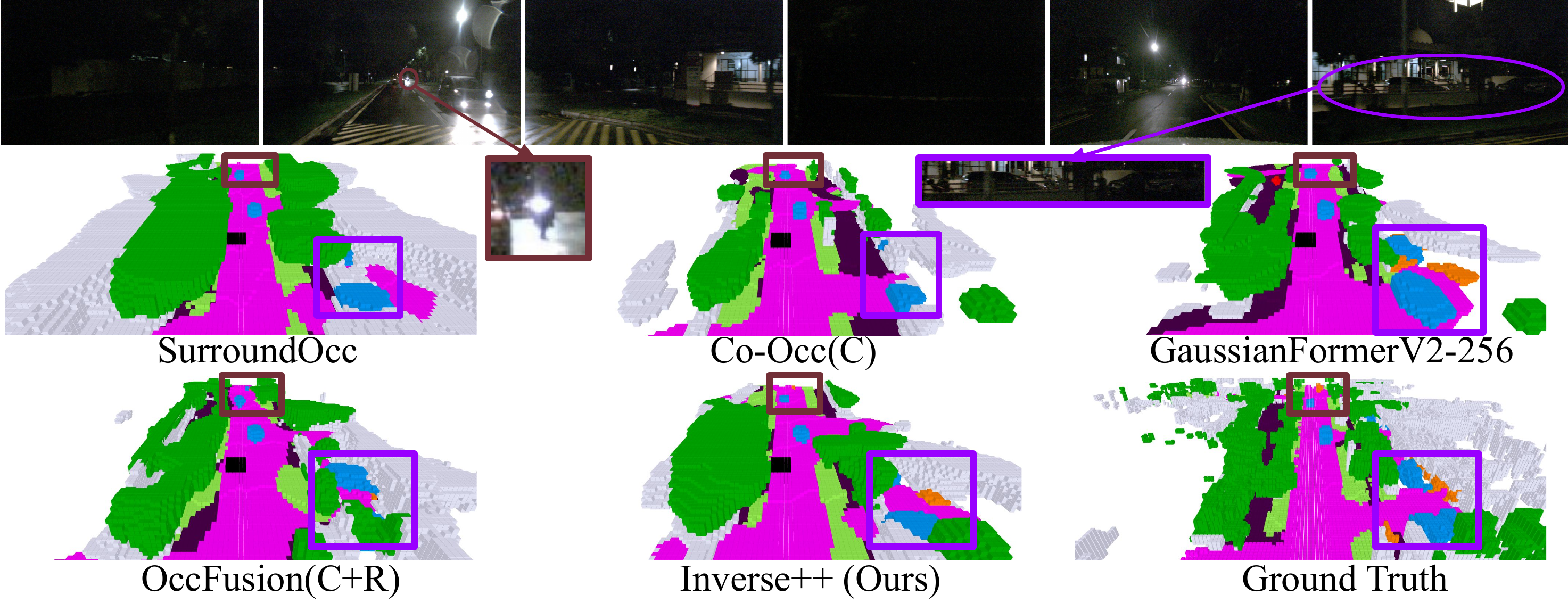}
\end{subfigure}
\caption{Qualitative results for daytime, rainy, and nighttime scenarios displayed in the upper, middle, and bottom sections, respectively. \textbf{Better viewed when zoomed in.} Notion of modality: Camera (C), Lidar (L), Radar (R).}
\label{qualitative}
\end{figure*}
We conducted qualitative analysis by generating visualizations of recent SOTA algorithms and comparing them with the prediction results from our work. The comprehensive visualization outcomes are depicted in Figure \ref{qualitative}. The top section illustrates the prediction results for the daytime scenario, the middle section displays the predictions for the rainy scenario, and the bottom section showcases the nighttime scenario results. A few circles with different colours signify the primary challenging area in the scene, while corresponding rectangles highlight the principal disparity in each prediction result for each algorithm.

In the daytime scenario, as shown in Figure \ref{qualitative} upper, all algorithms successfully detect the remote walking pedestrians (Highlighted in the green rectangle) on the sidewalk due to good lighting conditions and no occlusion. However, for the severely occluded front vehicle, which is highlighted with a dark red rectangle in the image, all SOTA algorithms, including OccFusion(C+R), which is a multi-modality fusion approach, failed to detect that vehicle except our algorithm, thanks to the extra 3D supervision signal applied on the intermediate features during training.

In the rainy scenario depicted in the middle of Figure \ref{qualitative}, a few pedestrians experience severe occlusion by building walls, trees, or vehicles parked along the roadside, presenting a challenging scenario for the algorithm. In this context, our algorithm leverages an additional 3D supervision signal from the auxiliary 3D object detection branch, enabling it to successfully detect all pedestrians—a feat unmatched by any other algorithm.

In the nighttime scenario, due to the nature of the camera, which is sensitive to the ambient lighting condition, all vision-centric approaches perform poorly in this scenario, as shown in Figure \ref{qualitative} bottom. Remarkably, our algorithm excels in detecting dynamic objects within the scene. Notably, we stand out as the sole algorithm capable of successfully identifying the motorcycle (highlighted in the dark red rectangle in the image) despite its considerable distance from the ego vehicle on the road.

\subsection{Model Efficiency}
\begin{table}[htbp]
  \centering
  \begin{adjustbox}{width=0.9\columnwidth}
  \begin{tabular}{c|cc}
    \toprule
    {Method} & {Latency (s) ($\downarrow$)} & {Memory (GB) ($\downarrow$)} \\
    \midrule
    NeWCRFs \cite{newcrfs} & 1.07 & 14.5 \\
    MonoScene \cite{monoscene} & 0.87 & 20.3 \\
    Adabins \cite{adabins} & 0.75 & 15.5 \\
    SurroundDepth \cite{surrounddepth} & 0.73 & 12.4\\
    SurroundOcc \cite{surroundocc} & 0.34 & 5.9 \\
    TPVFormer \cite{tpvformer} & 0.32 & 5.1 \\
    InverseMatrixVT3D \cite{ming2024inversematrixvt3d} & 0.32 & 4.82 \\
    BEVFormer \cite{bevformer} & \textbf{0.31} & \textbf{4.5} \\
    \hline
    Inverse++ & 0.32 & 7.9 \\
    
    \bottomrule
  \end{tabular}
  \end{adjustbox}
  \caption{Model efficiency comparison of different methods. The experiments are performed on a single RTX 3090 using six multi-camera images. For input image resolution, all methods adopt $1600\times900$. $\downarrow$:the lower, the better.}
  \label{efficiency}
\end{table}
Table \ref{efficiency} presents a comparison of inference time and memory usage across various methods. The experiments were carried out on a single RTX 3090 using six surround-view images with a resolution of $1600\times900$. Our approach, which integrates an additional 3D object-detection auxiliary branch, results in higher memory consumption. However, the latency remains comparable to that of other algorithms.

\subsection{Ablation Study}
\subsubsection{Encoder-Decoder Structure}
We conduct an ablation study on the encoder-decoder architecture, and the results are shown in Table \ref{abla_encoder_decoder}. The findings validate the significance of both the encoder and decoder in enhancing model performance through the detailed refinement of features. The absence of either component leads to a performance degradation of 0.5\% to 1.3\%.
\begin{table}[h]
  \centering
  \begin{adjustbox}{width=0.65\columnwidth}
  \begin{tabular}{cc|cc}
    \toprule
    Encoder & Decoder & IoU ($\uparrow$) & mIoU ($\uparrow$) \\
    \midrule
     &  & 28.83 & 15.86 \\
     & \XSolidBrush & 28.30 & 14.53 \\
    \XSolidBrush &  & 28.48 & 15.31 \\
    \XSolidBrush & \XSolidBrush & 28.54 & 15.48 \\
    \bottomrule
  \end{tabular}
  \end{adjustbox}
  \caption{\small Ablation study on encoder-decoder structure. $\uparrow$:the higher, the better.}
  \label{abla_encoder_decoder}
\end{table}

\subsubsection{3D Object Detection Auxiliary Branch}
We conducted an ablation study on the components of the auxiliary 3D object detection branch, and the results of the experiments are summarized in Table \ref{abla_auxiliary_branch}. The results indicate that each submodule within the auxiliary 3D object detection branch improves the model's overall performance by 0.9\% to 1.9\%. Notably, the visual cross-attention and 3D feature volume cross-attention modules make the most significant contributions to the model's overall performance.
\begin{table}[h]
  \centering
  \begin{adjustbox}{width=0.9\columnwidth}
  \begin{tabular}{cccc|cc}
    \toprule
    Self-Atten & Visual CA & BEV CA & 3D volume CA & IoU ($\uparrow$) & mIoU ($\uparrow$)\\
    \midrule
     &  &  &  & 28.43 & 15.86 \\
    \XSolidBrush &  &  &  & 28.15 & 14.98 \\
     & \XSolidBrush &  &  & 27.04 & 13.47 \\
     &  & \XSolidBrush &  & 27.83 & 14.56 \\
     &  &  & \XSolidBrush & 27.61 & 13.97 \\
    \bottomrule
  \end{tabular}
  \end{adjustbox}
  \caption{\small Ablation study on components of auxiliary 3D object detection branch. Self-Atten: self-attention module, Visual CA: visual cross-attention module, BEV CA: BEV feature cross-attention module, 3D volume CA: 3D feature volume cross-attention module.$\uparrow$:the higher, the better.}
  \label{abla_auxiliary_branch}
\end{table}

Furthermore, the impact of the auxiliary branch and encoder-decoder structure on detecting small and dynamic objects, including VRUs, on the road is demonstrated in Table \ref{abla_a3d-ed}. The experimental results highlight that our proposed modules substantially enhance the model's performance in detecting small and dynamic objects and excel in detecting VRUs, such as bicycles, motorcycles and pedestrians.
\begin{table}[h]
  \centering
  \begin{adjustbox}{width=0.9\columnwidth}
  \begin{tabular}{c|cccccc}
    \toprule
    A3D-ED & \rotatebox{90}{\textcolor{bus}{$\bullet$} bus} & \rotatebox{90}{\textcolor{car}{$\bullet$} car} & \rotatebox{90}{\textcolor{bicycle}{$\bullet$} bicycle} & \rotatebox{90}{\textcolor{motorcycle}{$\bullet$} motorcycle} & \rotatebox{90}{\textcolor{pedestrian}{$\bullet$} pedestrian} & \rotatebox{90}{\textcolor{cone}{$\bullet$} traffic cone}\\
    \midrule
    \XSolidBrush & 26.30 & 29.11 & 12.46 & 15.74 & 14.78 & 11.38 \\
    \Checkmark & \textbf{28.40} & \textbf{31.37} & \textbf{13.27} & \textbf{17.76} & \textbf{15.39} & \textbf{13.49} \\
     & (\textcolor{green}{+2.10}) & (\textcolor{green}{+2.26}) & (\textcolor{green}{+0.81}) & (\textcolor{green}{+2.02}) & (\textcolor{green}{+0.61}) & (\textcolor{green}{+2.11}) \\
    \bottomrule
  \end{tabular}
  \end{adjustbox}
  \caption{\small The ablation study investigates the influence of A3D-ED on the detection of VRU. A3D-ED refers to the auxiliary 3D object detection branch and encoder-decoder.}
  \label{abla_a3d-ed}
\end{table}

\section{Conclusion}\label{conclusion}
In this paper, we propose Inverse++, a vision-centric 3D semantic occupancy prediction method that assists with 3D object detection. Our approach first augments the previous InverseMatrixVT3D work with a U-Net-like encoder-decoder structure to further enhance its feature refinement capability. Then, an auxiliary 3D object detection branch is incorporated to introduce an extra 3D supervision signal, which is applied to the intermediated features to enhance the model's capability in capturing small dynamic objects. Unlike other SOTA algorithms, which depend on either a single 3D supervision signal or a combination of one 3D supervision signal and an additional 2D/2.5D supervision signal to improve the overall performance of the model, our approach utilizes two 3D supervision signals in the training phase. Extensive experiments conducted on the nuScenes datasets, including challenging rainy and nighttime scenarios, demonstrate that our method not only excels in its effectiveness but also achieves the best performance in detecting VRU for autonomous driving and road safety.

\bibliographystyle{IEEEtran}
\bibliography{mybibtex}

\end{document}

%% file: teaser_fig.tex
\begin{center}
    \captionsetup{type=figure}
    \includegraphics[trim={0cm 21.3cm 0 0},clip,width=\linewidth]{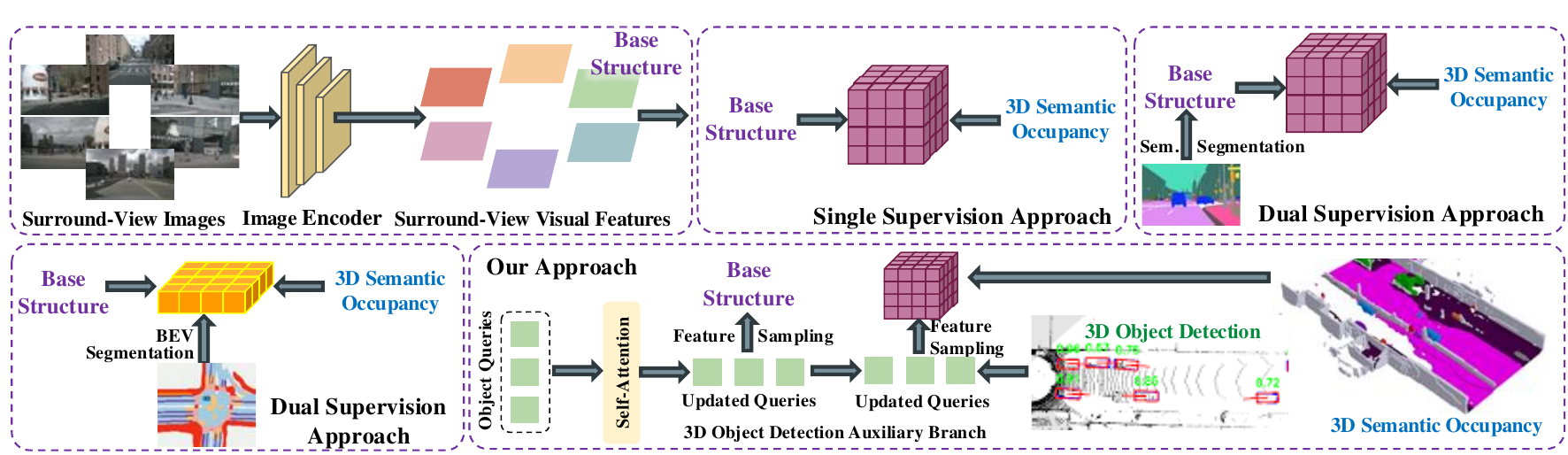}
    \captionof{figure}{\small The pipeline of four approaches: the single supervision signal-based, the dual supervision signal-based, and our proposed approach. To introduce an additional 3D supervision signal during training, we incorporate a 3D object detection auxiliary branch.}
    \label{teaser}
\end{center}